\documentclass[11pt,letterpaper]{article}
\usepackage{emnlp2017}
\usepackage{times}
\usepackage{latexsym}

\emnlpfinalcopy

\def\emnlppaperid{***}
\usepackage{url}
\usepackage{multirow}
\usepackage{times}
\usepackage{url}
\usepackage{rtrees}
\usepackage{pstricks}
\usepackage{amsmath}
\usepackage{amsfonts}
\usepackage{amssymb}
\usepackage{graphicx}
\usepackage[caption=false,font=footnotesize]{subfig}
\usepackage{cleveref}
\usepackage[T1]{fontenc}

\newcommand{\ignore}[1]{}

\newcommand{\babble}[0]{\textsc{babble}}
\newcommand{\memnn}[0]{\textsc{memn2n}}
\newcommand{\memnns}[0]{\textsc{memn2n}s}

\usepackage{url}
\usepackage{dsttr}
\begin{document}
%


\title{Bootstrapping incremental dialogue systems from minimal data: the generalisation power of dialogue grammars}



 

\author{Arash Eshghi \\
  Interaction Lab   \\
 Heriot-Watt University  \\
  {\tt  a.eshghi@hw.ac.uk} \\ \And
   Igor Shalyminov\\
 Interaction Lab   \\
 Heriot-Watt University  \\
  {\tt is33@hw.ac.uk} \\ \And
  Oliver Lemon \\
 Interaction Lab   \\
 Heriot-Watt University  \\
  {\tt o.lemon@hw.ac.uk}}


\maketitle

\begin{abstract}
We investigate an end-to-end method for automatically inducing task-based dialogue systems from small amounts  of unannotated dialogue data. It combines an incremental semantic grammar  - Dynamic Syntax and Type Theory with Records (DS-TTR) - with Reinforcement Learning (RL), where language generation and dialogue management are a joint  decision problem. The systems thus produced are {\it incremental}: dialogues are processed word-by-word, shown previously to be essential in supporting natural, spontaneous dialogue. We hypothesised that the rich linguistic knowledge within the grammar should enable a combinatorially large number of dialogue variations to be processed, even when trained on very few dialogues. Our experiments show that our model can process 74\% of the Facebook AI bAbI dataset even when trained on only 0.13\% of the data (5 dialogues). 
It can \textit{in addition} process 65\% of bAbI+, a corpus\footnote{Dataset available at \url{https://bit.ly/babi_plus}} we created by systematically adding incremental dialogue phenomena such as restarts and self-corrections to bAbI. We compare our model with a state-of-the-art retrieval model, \memnn\ \cite{babi}.
We find that, in terms of semantic accuracy, \memnn\ shows very poor robustness to the bAbI+ transformations even when trained on the full bAbI dataset. 



\end{abstract}



\section{Introduction}\label{tab:introduction}


There are currently several key problems for the practical data-driven  (rather than hand-crafted) development of task-oriented dialogue systems, among them: (1) large amounts of dialogue data are needed, i.e.\ thousands of examples in a   domain; (2) this data is usually required to be annotated with task-specific semantic/pragmatic information for the domain (e.g.\ various dialogue act schemes); and (3) the resulting systems are generally turn-based, and so do not support natural spontaneous dialogue which is processed {\it incrementally}, word-by-word, with many characteristic phenomena that arise from this incrementality.

In overcoming issue (2),  a recent advance made in research on (non-task)  chat dialogues has been the development of so-called ``end-to-end" systems, in  which all components are trained from textual dialogue examples, e.g.\ \cite{sordoni,vinyals}.
However, as Bordes and Weston  \shortcite{babi} argued, these end-to-end methods may not transfer well to task-based settings (where the user is trying to achieve a domain goal, such as booking a flight or finding a restaurant, resulting in an API call).  
Bordes and Weston  \shortcite{babi} then presented an end-to-end method using Memory Networks (\memnns), which achieves 100\% performance on a test-set of 1000 dialogues, after being trained on 1000 dialogues. This method processes dialogues turn-by-turn, and so does not have the advantages of more natural incremental systems \cite{aistincremental,Skantze.Hjalmarsson10}; nor does it really perform language generation, rather it's based on a retrieval model that selects from a set of candidate system responses seen in the data.

This paper investigates an approach to these challenges - dubbed \babble\ - using an incremental, semantic parser and generator for dialogue \cite{Eshghi.etal11,Eshghi15}, based around the Dynamic Syntax grammar formalism (DS, \newcite{Kempson.etal01,Cann.etal05a}).



Our advance in this paper, for end-to-end systems, is therefore twofold: (a) the \babble\ method overcomes the requirement for large amounts of dialogue data (i.e.\ 1000s of dialogues in a domain); (b) resulting systems are word-by-word incremental, in both parsing, generation and dialogue management. We show that using only 5 example dialogues from the bAbI, Task 1 dataset (i.e.\ 0.13\% of the training data used by  \cite{babi}) \babble\ can automatically induce dialogue systems which process 74\% of the bAbI testset in an incremental manner. We then introduce an extended incremental version of the bAbI dataset, which we call bAbI+ (see section \ref{babi+}), which adds some characteristic incremental phenomena - such as mid-utterance self-corrections - to the bAbI dialogues (this new dataset is freely available).
Using this, we demonstrate that the \babble\ system can in addition generalise to, and process 65\% of the bAbI+ dataset, still when trained \emph{only on 5 dialogues from bAbI}. We compare this method to \cite{babi}'s \memnn, which, in terms of semantic accuracy (reflected in how well \texttt{api-call}s are predicted at the end of bAbI Task 1), shows very poor robustness to the bAbI+ transformations, even when it is trained on \emph{the full bAbI dataset}.


This overall method is portable to other task-based domains. Furthermore, as we use a semantic parser, the semantic/contextual representations of the dialogue can be used directly for large-scale inference, required in more complex tasks (e.g. interactive QA and search).

\ignore{
Dialogue is domain-specific, in that the communicative import of utterances is severely underdetermined in the absence of a specific domain of language use. Even within a simple domain, there's a lot of variation in language use that does not ultimately affect the overall communicative goal of the dialogue. For example,}

\ignore{ 
\subsection{Pragmatic synonymy and dialogue variation}
 The dialogues in Fig.~\ref{variation} all lead to a context in which the user wants to buy a phone by LG. These dialogues can be said to be \emph{pragmatically synonymous} for this domain.

To capture such synonymy,   developers of task-oriented dialogue systems have relied on hand-crafted, and manual Dialogue Act (DA)  annotations\footnote{Here we use the term ``dialogue act'' to encompass the whole semantic representation used, ie.\ standard dialogue acts such as ``inform'' together with content such as ``product-type=LG''.} which, in effect, provide a model of a particular domain. But these are very expensive to design and annotate, and also lead to systems which lack generality and do not easily scale. In standard task-based dialogue system development paradigms,  DAs form a bottleneck representation between language understanding  and Dialogue Management (DM), and between DM and Natural Language Generation (NLG). In addition, from a machine-learning point of view  DA representations may either
under- or over-estimate the features required for learning
good DM and/or NLG policies for a domain. When using Reinforcement Learning methods, as in e.g. Young et. al \shortcite{Young.etal13};Rieser et al. \shortcite{Rieser.etal11}, the MDP state space must also be hand-crafted, and the mapping from DAs to dialogue states normally must be explicitly defined (c.f.\ Thomson \& Young \shortcite{Thomson.Young10}). 

In this paper, we present an implemented, ``end-to-end" method for data-driven learning of fully incremental dialogue systems using small amounts of   dialogue transcripts, without the use of dialogue act annotations. Incremental dialogue systems (i.e.\ processing word-by-word instead of at utterance/turn boundaries) have previously  been empirically shown to be beneficial and more natural for users \cite{aistincremental,Skantze.Hjalmarsson10}. 

Our method combines an incremental, semantic grammar formalism - Dynamic Syntax (DS) - with Reinforcement Learning (RL) for incremental word selection, where dialogue management and language generation are treated as one and the same decision/optimisation problem, and where {\it the corresponding MDP is automatically constructed}. Using our implemented system, we demonstrate that using an incremental model of dialogue such as DS, one can generalise from very small amounts of raw dialogue data, to a combinatorially large space of interactional variations, including phenomena such as question-answer pairs, over-answering, self- and other-corrections, split-utterances, and clarification interaction, when many of these are not even observed in the original data (see section \ref{evaluation}).

%
}  

\subsection{Dimensions of Pragmatic Synonymy}\label{synonymy}
There are two important dimensions along which dialogues can vary, but nevertheless, lead to identical contexts: interactional, and lexical. Interactional synonymy is analogous to syntactic synonymy - when two distinct sentences are parsed to identical logical forms - except that it occurs not only at the level of a single sentence, but at the dialogue or discourse level. Fig.~\ref{variation} shows examples of interactional variants that lead to very similar final contexts, in this case, that the user wants to buy an LG phone. These dialogues can be said to be \emph{pragmatically synonymous} for this domain. Arguably, a good computational model of dialogue processing, and interactional dynamics should be able to capture this synonymy. 

Lexical synonymy relations, on the other hand, hold among utterances, or dialogues, when different words (or sequences of words) express meanings that are sufficiently similar in a particular domain.
What is striking about lexical synonymy relations is that unlike syntactic/interactional ones, they can often  break down when one moves to another domain: lexical synonymy relations are domain specific. 


Eshghi \& Lemon \shortcite{Eshghi.Lemon14} developed a method similar in spirit to Kwiatkowski et al. \shortcite{Kwiatkowski.etal13} for capturing lexical synonymy relations by creating clusters of semantic representations based on observations that they give rise to similar or identical extra-linguistic actions observed within a domain (e.g.\ a data-base query, a flight booking, or any API call). Distributional methods could also be used for this purpose (see e.g. Lewis \& Steedman \shortcite{Lewis.Steedman13}). In general, this kind of clustering is achieved when the domain-general semantics resulting from semantic parsing is grounded in a particular domain.

We note that while interactional synonymy relations in dialogue should be accounted for by semantic grammars or formal models of dialogue structure (such as DS-TTR \cite{Eshghi.etal12}, or KoS \cite{Ginzburg12}), lexical synonymy relations have to be learned from data. 


\begin{figure*}[ht]
\begin{footnotesize}
\begin{tabular}{|r|l|l|l|}\hline
\parbox[t]{3mm}{\multirow{3}{*}{\rotatebox[origin=c]{90}{Interactional Variation}}}&
\begin{tabular}{ll}
USR: & I would like an LG laptop\\
&sorry uhm phone\\
SYS: & okay.
\end{tabular}&
\begin{tabular}{ll}
USR: & I would like a\\ 
&phone by LG.\\
SYS: & sorry a what?\\
USR: & a phone by LG.\\
SYS: & okay.
\end{tabular}&
\begin{tabular}{ll}
SYS: & what would you like?\\
USR: & an LG phone\\
SYS: & okay.
\end{tabular}\\\cline{2-4}
&
\begin{tabular}{ll}
SYS: & what would you like?\\
USR: & a phone\\
SYS: & by which brand?\\
USR: & LG\\
SYS: & okay
\end{tabular}&
\begin{tabular}{ll}
SYS: & you'd like a ...?\\
USR: & a phone\\
SYS: & by what brand?\\
USR: & LG.\\
SYS: & okay
\end{tabular}&
\begin{tabular}{ll}
SYS: & so would you like a computer?\\
USR: & no, a phone.\\
SYS: & okay. by which brand?\\
USR: & LG.\\
SYS: & okay.
\end{tabular}\\&&&\\\hline



\end{tabular}
\end{footnotesize}
\caption{Some Interactional Variations in a Shopping Domain}\label{variation}\vspace{-0.5cm}
\end{figure*}

\section{Why a grammar-based approach?} \label{sec:related_work}



Recent end-to-end data-driven machine learning approaches treat dialogue as a sequence-to-sequence generation problem, and train their models from large datasets e.g.\ \cite{sordoni,Wen.etal16a,Wen.etal16b,vinyals}. The systems resulting from these types of approach are in principle able to handle variations/patterns that they have encountered (sufficiently often) in the training data, but not beyond.

This large-data constraint is problematic for developers but  is also  strange when we consider the structural knowledge that we have about language and dialogue that can be encoded in grammars and computational models of interaction. Indeed, it is often stated that for humans to learn how to perform adequately in a domain, one example is enough from which to learn (e.g.\ Li et. al \shortcite{DBLP:journals/pami/Fei-FeiFP06}). 


Furthermore, as these systems do not parse to logical forms, they do not allow for explicit inference, which further limits their application. 

We therefore develop a method combining learning from data with  an incremental semantic grammar of dialogue that is able to generalise from   small number of observations in a domain -- in fact even from just a few examples of  successful dialogues -- to a large range of interactional and syntactic variations, including everyday natural incremental phenomena. 

\section{Inducing Dialogue Systems}\label{model}

Our overall method involves incrementally parsing dialogues, and encoding the resulting semantics as state vectors in a Markov Decision Process (MDP), which is then used for Reinforcement Learning (RL) of word-level actions for system output (i.e.\ a combined incremental DM and NLG module for the resulting dialogue system). 

 \subsection{Dynamic Syntax and Type Theory with Records (DS-TTR)}\label{dsttr}
\begin{figure*}
\includegraphics[width=\linewidth]{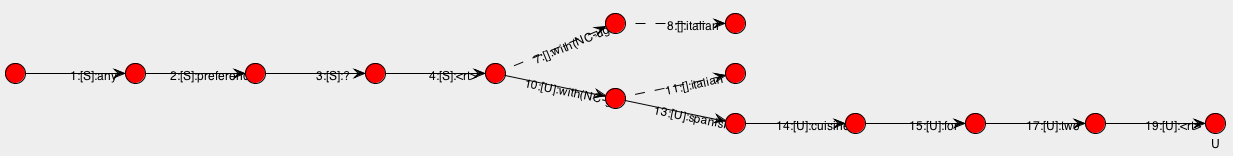}
\caption{DS-TTR: Incremental Parsing of self-corrections and restarts}\label{fig:dag}
\end{figure*}

\textbf{Dynamic Syntax (DS)} is an action-based, word-by-word incremental and semantic grammar formalism \cite{Kempson.etal01,Cann.etal05a}, especially suited  to the highly fragmentary and context-dependent nature of dialogue. In DS, words are conditional actions - semantic updates; and dialogue is modelled as the interactive and incremental construction of contextual and semantic representations \cite{Eshghi.etal15} - see Fig.~\ref{fig:subtype}. The contextual representations afforded by DS are of the fine-grained semantic content that is jointly negotiated/agreed upon by the interlocutors, as a result of processing questions and answers, clarification interaction, acceptances, self-/other-corrections, restarts, and other characteristic incremental phenomena in dialogue - see~\ref{fig:dag} for a sketch of how self-corrections and restarts are processed via a backtrack and search mechanism over the parse search graph (see \newcite{Hough11, Hough.Purver14, Eshghi.etal15} for details of the model, and how this parse search graph is effectively the context of the conversation). Generation/linearisation in DS is defined using trial-and-error parsing (see Section~\ref{sec:babble}, with the provision of \emph{a generation goal}, viz. the semantics of the utterance to be generated. Generation thus proceeds, just as with parsing, on a word-by-word basis (see \newcite{Purver.etal14,Hough15} for details).
The upshot of this is that using DS, we can not only track the semantic content of some current turn as it is being constructed (parsed or generated) word-by-word, but also the context of the conversation as whole, with the latter also encoding the grounded/agreed content of the conversation (see e.g. Fig. \ref{fig:encoding}, and see Eshghi et al. (2015); Purver et al. \shortcite{Purver.etal10} for details). Crucially for our model below, the inherent incrementality of DS-TTR together with the word-level, as well as cross-turn, parsing constraints it provides, enables the word-by-word exploration  of the space of grammatical dialogues, and the semantic and contextual representations that result from them. 
 
\begin{figure*}[!ht]
\begin{footnotesize}
\centerline{
\begin{tabular}{c@{$\quad\mapsto\quad$}c@{$\quad\mapsto\quad$}c@{$\quad\mapsto\quad$}c}
$\ttrnode{}{event&e_s\\p1_{=today(event)}&t}$ &
$\ttrnode{}{event_{=arrive}&e_s\\p1_{=today(event)}&t\\p2_{=pres(event)}&t\\x_{=robin}&e\\p3_{=subj(event,x)}&t}$ &
$\ttrnode{}{event_{=arrive}&e_s\\p1_{=today(event)}&t\\p2_{=pres(event)}&t\\x_{=robin}&e\\p3_{=subj(event,x)}&t\\x1&e\\p3_{=from(event,x1)}&t}$ &
$\ttrnode{}{event_{=arrive}&e_s\\p1_{=today(event)}&t\\p2_{=pres(event)}&t\\x_{=robin}&e\\p_{=subj(event,x)}&t\\x1_{=Sweden}&e\\p3_{=from(event,x1)}&t}$
\\[5em]
\emph{``A: Today''} & \emph{``..Robin arrives''} & \emph{``B: from?''} & \emph{``A: Sweden''}
\end{tabular}
}
\end{footnotesize}
\caption{Incremental parsing using DS-TTR}\label{fig:subtype}

\end{figure*}

\paragraph{Type Theory with Records (TTR)} is an extension of standard type theory shown to be
useful in semantics and dialogue modelling \cite{Cooper05,Ginzburg12}. To accommodate dialogue processing, and allow for richer representations of the dialogue context recent work has integrated DS and the TTR framework to replace the logical formalism in which meanings are expressed \cite{Purver.etal10,Purver.etal11,Eshghi.etal12}. In TTR, logical forms are specified as \emph{record types} (RTs), sequences of \emph{fields} of the form $\smttrnode{}{l&T}$ containing a label $l$ and a type
$T$. RTs can be witnessed (i.e.~judged as true) by \emph{records} of that type, where a record is a sequence of label-value pairs $\smttrrec{}{l&v}$, and $\smttrrec{}{l&v}$ is of type $\smttrnode{}{l&T}$ just in case $v$ is of type $T$ (see Fig.\ \ref{fig:subtype} for example record types).



Importantly for us here, the standard
\textit{subtype} relation $\subtype$ can be defined for record types: $R_1 \subtype
R_2$ if for all fields $\smttrnode{}{l & T_2}$ in $R_2$, $R_1$ contains
$\smttrnode{}{l & T_1}$ where $T_1 \subtype T_2$. 
A record type can thus be indefinitely extended, and is therefore always underspecified by definition. This allows for incrementally growing meanings to be expressed in a natural way as more words are parsed or generated in turn. In addition, as will become clear below, this subtype checking operation is the key mechanism used in our system below for feature checking.


 \subsection{Overall Method: \babble}
 \label{sec:babble}
In this section we describe our method for combining incremental dialogue parsing with Reinforcement Learning for Dialogue Management (DM) and Natural Language Generation (NLG) where these are treated as a joint decision/optimisation problem.

We start with two resources: 
a) a DS-TTR
parser  $DS$ (either learned from data \cite{Eshghi.etal13a}, or constructed by hand), for incremental language processing, but also, more generally, for tracking the context of the dialogue using Eshghi et al.'s model of feedback \cite{Eshghi.etal15,Eshghi15,Eshghi.etal11}; 
b) a set
$D$ of transcribed successful  dialogues in the target  domain. 

We perform the following steps overall to induce a fully incremental dialogue system from $D$:

\begin{enumerate}
\setlength{\itemsep}{-2pt} 
\item Automatically induce the MDP state space, $S$, and the dialogue goal, $G_D$, from $D$;

\item Automatically define the state encoding function $F: C \rightarrow S$; where $s \in S$ is a (binary) state vector, designed to extract from the current context of the dialogue, the semantic features observed in the example dialogues $D$; and $c \in C$ is a DS context, viz.\ a pair of TTR Record Types: $\langle c_{p}, c_{g}\rangle$, where $c_{p}$ is the content of the current, \textit{PENDING} clause as it is being constructed, but not necessarily fully grounded yet; and $c_{g}$ is the content already jointly built and \textit{GROUNDED} by the interlocutors (loosely following the DGB model of \cite{Ginzburg12}).

\item Define the MDP action set as the $DS$ lexicon $L$ (i.e.\ actions are words);

\item Define the reward function $R$ as reaching $G_D$, while minimising dialogue length. 

\end{enumerate}

We then solve the generated MDP using Reinforcement Learning, with a standard Q-learning method, implemented using BURLAP \cite{burlap}: train a policy $\pi : S \rightarrow L$, where $L$ is the DS Lexicon, and $S$ the state space induced using $F$. The system is trained in interaction with a (semantic) simulated user, also automatically built from the dialogue data and described in the next section.



\begin{figure*}[ht]
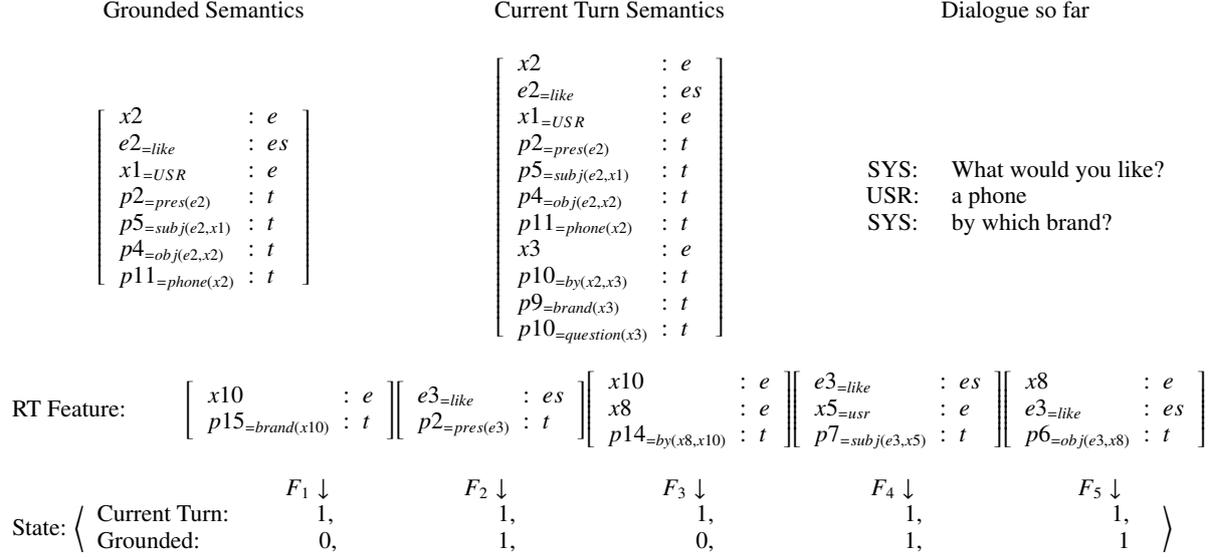

\centering
\begin{small}
\begin{tabular}{llcccccl}
\multicolumn{3}{c}{Grounded Semantics}&\multicolumn{2}{c}{Current Turn Semantics}&\multicolumn{2}{c}{Dialogue so far}\\\\
\multicolumn{3}{c}{$\ttrnode{}{x2&e
\\e2_{=like}&es\\x1_{=USR}&e\\p2_{=pres(e2)}&t\\p5_{=subj(e2, x1)}&t\\p4_{=obj(e2, x2)}&t\\p11_{=phone(x2)}&t}$}&\multicolumn{2}{c}{$\ttrnode{}{x2&e
\\e2_{=like}&es\\x1_{=USR}&e\\p2_{=pres(e2)}&t\\p5_{=subj(e2, x1)}&t\\p4_{=obj(e2, x2)}&t\\p11_{=phone(x2)}&t\\x3&e\\p10_{=by(x2, x3)}&t\\p9_{=brand(x3)}&t\\p10_{=question(x3)}&t}$}&\multicolumn{2}{c}{\begin{tabular}{cl}
SYS: & What would you like?\\
USR: & a phone\\
SYS: & by which brand?\\
\end{tabular}}\\\\
RT Feature:&&\hspace{-1cm}$\ttrnode{}{x10&e\\
p15_{=brand(x10)}&t}$&\hspace{-0.5cm}
$\ttrnode{}{e3_{=like}&es\\
p2_{=pres(e3)}&t}$&\hspace{-0.5cm}
$\ttrnode{}{x10&e\\
x8&e\\
p14_{=by(x8, x10)}&t}$&\hspace{-0.5cm}
$\ttrnode{}{e3_{=like}&es\\
x5_{=usr}&e\\
p7_{=subj(e3, x5)}&t}$&\hspace{-0.5cm}
$\ttrnode{}{x8&e\\
e3_{=like}&es\\
p6_{=obj(e3, x8)}&t}$\\\\
&&\hspace{-0.55cm}$F_1\downarrow$&\hspace{-0.55cm}$F_2\downarrow$&\hspace{-0.55cm}$F_3\downarrow$&\hspace{-0.55cm}$F_4\downarrow$&\hspace{-0.55cm}$F_5\downarrow$\\
\multirow{2}{*}{State: $\bigg\langle$}&\hspace{-0.8cm}Current Turn:& \hspace{-0.17cm} 1, & 1,& 1,& 1,& 1,&\multirow{2}{*}{\hspace{-1cm}$\bigg\rangle$}\\
&\hspace{-0.8cm}Grounded:&  0,& 1,& 0,& 1,& \hspace{0.1cm}1
\end{tabular}
\end{small}
\caption{Semantics to MDP state encoding with RT features}\label{fig:encoding}\vspace{-0.5cm}
\end{figure*}

\paragraph{The state encoding function, $F$}
As shown in figure \ref{fig:encoding} the MDP state is a binary vector of size $2\times | \Phi |$, i.e.\ twice the number of the RT features. The first half of the state vector contains the grounded features (i.e.\ agreed by the participants)    $\phi_i$, while the second half contains the current semantics being incrementally built in the current dialogue utterance. Formally:\\
$s = \langle F_1(c_{p}), \ldots, F_m(c_{p}), F_1(c_{g}), \ldots, F_m(c_{g})\rangle$; \newline
where $F_i(c)=$ 
    1 if  $c\subtype\phi_i$, and 
    0            otherwise.
(Recall that $\subtype$  is the RT subtype relation).

\subsubsection{Semantic User Simulation}\label{usersim}

The simulator is in charge of two key tasks during training: (1) generating user turns in the right dialogue contexts; and (2) word-by-word monitoring of the utterance so far generated by the system during exploration (i.e.\ babbling grammatical word sequences) by the system. To exploit (and evaluate) the full generalisation properties of the $DS$ dialogue model, both (1) and (2) use the full machinery of the $DS$ parser, as well the state encoding function $F$, described above. They are thus performed based on the  \textit{semantic context} of the dialogue so far, as tracked by $DS$ (rather than, e.g.\ being based on string or template matching). Since this includes not just the semantic features of the current turn, but also of \emph{the history of the conversation}, our simulator respects the turn orderings encountered in the data, i.e. it is sensitive to the order in which information is gathered from the user.

The rules required for (1) \& (2) are extracted \textit{automatically} from the raw dialogue data, $D$, using $DS$ and $F$. The dialogues in $D$ are parsed and encoded using $F$ incrementally. For (1), all the states that trigger the user into action, $s_{i}=F(c)$ -- where $c$ is a DS context -- immediately prior to any user turn are recorded, and mapped to what the user ends up saying in those contexts - for more than one training dialogue there may be more than one candidate (in the same context/state). The rules thus extracted will be of the form:\\ $s_{trig} \rightarrow \{u_1, \ldots, u_n\}$, where $u_i$ are user turns.

Now note that the $s_i$'s prior to the user turns also immediately follow system turns. And thus to perform (2), i.e.\ to monitor the system's behaviour during training, we only need to check further that the current state resulting from processing a word generated by the system, subsumes - is extendible to - one of the $s_i$. We perform this through a simple bitmask operation (recall that the states are binary). The simulation can thus semantically identify erroneous/out-of-domain actions (words) by the system. It would then terminate the learning episode and penalise the system immediately, aiding speed of training significantly.

\section{Evaluation}\label{evaluation}
We have so far induced two prototype dialogue systems, one in an `electronics shopping' domain (see \newcite{Kalatzis.etal16} and Fig.~\ref{variation}) and another in a `restaurant-search' domain, showing that fully incremental dialogue systems can be automatically induced  from small amounts of unannotated dialogue transcripts \cite{Kalatzis.etal16, Eshghi.etal17} - in this case both systems were bootstrapped from a {\it single} successful example dialogue. We are in the process of evaluating these systems with real users. 

In this paper, however, our focus is not on building dialogue systems \textit{per se}, but on: \textbf{(1)} studying and quantifying the interactional and structural generalisation power of the DS-TTR grammar formalism (see Section~\ref{sec:related_work}), and that of symbolic, grammar-based approaches to language processing more generally. We focus here on specific dialogue phenomena, such as mid-sentence self-corrections, hesitations, and restarts (see below);
\textbf{(2)} doing the same for Bordes and Weston's \shortcite{babi} state-of-the-art, bottom up response retrieval model, without use of linguistic knowledge of any form; and
\textbf{(3)} comparing (1) and (2).

In order to test and quantify the interactional and structural generalisation power/robustness of the two models, \babble\ and \memnn, we need contrasting dialogue data-sets that control for interactional vs.\ lexical variations in the input dialogues. Furthermore, to make our results comparable to the existing approach of Bordes and Weston  \shortcite{babi}, we need to use the same dataset that they have used. We therefore use  Facebook AI Research's bAbI dialogue tasks dataset \cite{babi}. These are goal-oriented dialogues in the domain of restaurant search. Here we  tackle Task 1, where in each dialogue the system asks the user about their preferences for the properties of a
restaurant, and each dialogue results in an API call which contains values of
each slot obtained. Other than the explicit API call notation, there are no
annotations in the data whatsoever. 

\subsection{The bAbI+ dataset}\label{babi+}
While containing some lexical variation, the original bAbI dialogues significantly lack interactional variation vital for natural real-life dialogue. In order to
obtain such variation while holding lexical variation constant, 
we created the bAbI+ dataset by systematically transforming the bAbI dialogues. 

bAbI+ is an  extension of the original bAbI Task 1 dialogues with everyday incremental dialogue phenomena (hesitations, restarts, and corrections -- see below).  While the original bAbI tasks 2---7 increase the user's goal complexity, modifications introduced in bAbI+ can be thought of as orthogonal to this: we instead increase the complexity of surface forms of dialogue utterances, while keeping every other aspect of the task fixed.

The variations introduced in bAbI+ are: \textbf{1. Hesitations}, e.g.\ as in ``we will be \texttt{uhm} eight''; \textbf{2. Restarts}, e.g.\ ``can you make a restaurant \texttt{uhm yeah can you make a restaurant} reservation for four people with french cuisine in a moderate price range''; and \textbf{3. Corrections} affecting task-specific information - both short-distance ones correcting one token, e.g.\ ``with french \texttt{oh no spanish} food'', and long-distance NP/PP-level corrections, e.g. ``with french food \texttt{uhm sorry with spanish food}''.

The phenomena above are mixed in probabilistically from the fixed sets of templates to the original data\footnote{See \url{https://github.com/ishalyminov/babi_tools}}. The modifications affect a total of \textbf{11336} utterances in the \textbf{3998} dialogues. Around \textbf{21\%} of user turns contain corrections, \textbf{40\%} hesitations, and \textbf{5\%} restarts (they are not mutually exclusive, so that an utterance can contain up to 3 modifications). Our modifications, with respect to corrections in particular, are more conservative than those observed in real-world data: 
\newcite{Hough15} reports that self-corrections appear in \textbf{20\%} of all turns of natural conversations from the British National Corpus, and in \textbf{40\%} of turns in the Map Task, a corpus of human-human goal-oriented dialogues. Here's part of an example dialogue in the bAbI+ corpus:

\begin{small}
\begin{tabular}{lp{6cm}}
\textbf{sys:}&	hello what can I help you with today?\\
\textbf{usr:}&	I'd like to book a uhm yeah I'd like to book a table in a expensive price range\\
\textbf{sys:}&	I'm on it. Any preference on a type of cuisine?\\
\textbf{usr:}&	with indian food no sorry with spanish food please\\
\end{tabular}
\end{small}

\subsection{Memory Network setup}
In all the experiments we describe below, we follow Bordes and Weston's setup  by using a \memnn\ (we took an open source Tensorflow implementation for bAbI QA tasks and modified it\footnote{See \url{https://github.com/ishalyminov/memn2n}} according to their setup~-- see details below).
In order to adapt the data for the \memnn, we transform the dialogues into \textit{<story, question, answer>} triplets. The number of triplets for a single dialogue is equal to the number of the system's turns, and in each triplet, the answer is the current system's turn, the question is the user's turn preceding it, and the story is a list of all the previous turns among both sides.

The \memnn\ hyperparameters are set as follows: \textbf{1} hop, and \textbf{128} as the size of embeddings; we train it for \textbf{100} epochs with a learning rate of \textbf{0.01} and a batch size of \textbf{8} ~-- in this we follow the best bAbI Task 1 setup reported by \cite{babi}.

\subsection{Testing the DS-TTR parser} Dynamic Syntax (DS) lexicons are learnable from data \cite{Eshghi.etal13a,Eshghi.etal13b}. But since the lexicon was induced from a corpus of child-directed utterances in this prior work, there were some constructions as well as individual words that it did not include\footnote{We are currently looking into applying Eshghi et al.'s \shortcite{Eshghi.etal13a} model to induce DS grammars from larger semantic corpora such as the Groningen Meaning Bank, leading to much more wide-coverage lexicons}. One of the authors therefore extended this induced grammar manually to cover the bAbI dataset, which, despite not being very diverse, contains a wide range of complex grammatical constructions, such as long sequences of prepositional phrases, adjuncts, short answers to yes/no and wh-questions, appositions of NPs, causative verbs etc.

We parsed all dialogues in the bAbI train and test sets, as well as on the bAbI+ corpus word-by-word, including both user and system utterances, in context. The grammar  parses 100\% of the dialogues, i.e.\ it does not fail on any word in any of the dialogues. We assess the semantic accuracy of the parser on bAbI \& bAbI+ using the dialogue-final \texttt{api-call}s in section \ref{sec:sem-acc} below.

\subsection{Experiment 1: Generalisation from small data}\label{sec:comparison}

We have now set out all we need to perform the first experiment.
Our aim here is to assess the generalisation power that results from the grammar and our state encoding method
 (section \ref{dsttr}) - we dub our overall model \babble\ - 
and compare this to the state of the art results of \newcite{babi}. The method in \newcite{babi} is not generative, rather it is based on retrieval of system responses, based on the history of the dialogue up to that point. Therefore, for direct comparison, and for simplicity of exposition, we do the same here: we apply the method described for creating a user simulation (section \ref{usersim}), this time \textit{for the system side}, resulting in a `system simulation'. We then use this to predict a system response, by parsing and encoding the containing test dialogue up to the point immediately prior to the system turn. This results in a triggering state, $s_{trig}$, which is then used as the key to look up the system's response from the rules constructed as per section\ \ref{usersim}. The returned response is then parsed word-by-word as normal, and this same process continues for the rest of the dialogue. This method uses the full machinery of DS-TTR \& our state-encoding method - the \babble\ model - and will thus reflect the generalisation properties that we are interested in.

\paragraph{Cross-Validation} Since we are here interested in data efficiency and generalisation we use all the bAbI and bAbI+ data - the train, dev, and test sets - as follows: we train Bordes \& Weston's \memnn\ and \babble\ from 1-5 examples selected at random from \textit{the longest dialogues} in bAbI -- note \textbf{bAbI+ data is never used for training in these experiments}. This process is repeated across 10 folds. The models are then tested on sets of 1000 examples selected at random, in each fold. Both the training and test sets constructed in this way are kept constant in each fold across the \babble\ \& \memnn\ 
models. The test sets are selected either exclusively from bAbI or exclusively from bAbI+.

\subsubsection{Results: Predicting system turns}

Table~\ref{tab:results} shows per utterance accuracies for the \babble\ \& \memnn\ models. Per utterance accuracy is the percentage of all system turns in the test dialogues that were correctly predicted. The table shows that \babble\ can generalise to a remarkable 74\% of bAbI and 65\% of bAbI+ with only 5 input dialogues from bAbI. It also shows that \memnns\ can also generalise remarkably well. Although as discussed below, this result is misleading on its own as the \memnns\ are very poor at generating the final \texttt{api-call}s correctly on both the bAbI \& bAbI+ data, and are thus making too many semantic mistakes.

\begin{table*}
\begin{center}
\begin{tabular}{|c|c|c|c|c|c|}\hline
\# of training dialogues:&1&2&3&4&5\\\hline \hline 
\textbf{\babble\ on bAbI}&67.12&73.36&72.63&73.32&74.08\\\hline
\textbf{\memnn\ on bAbI}&2.77&59.15&70.94&71.68&72.6\\\hline
\textbf{\babble\ on bAbI+}&59.42&65.27&63.45&64.34&65.2\\\hline
\textbf{\memnn\ on bAbI+}&0.22&56.75&68.65&71.84&73.2\\\hline
\end{tabular}
\end{center} \caption{Mean per utterance accuracies (\%) for \memnn\ \& \babble\ models across the bAbI \& bAbI+ datasets (10 folds) }\label{tab:results}
\end{table*}

\subsection{Experiment 2: Semantic Accuracy}\label{sec:sem-acc}

The results from Experiment 1 on their own can be misleading, as correct prediction of system responses does not in general tell us enough about how well the models are interpreting the dialogues, or whether they are doing this with a sufficient level of granularity. To assess this, in this second experiment, we measure the semantic accuracy of each model by \emph{looking exclusively} at how accurately they predict the final \texttt{api-call}s in the bAbI \& bAbI+ datasets. For the \memnn\ model, we follow the same overall procedure as in the previous experiment: train on bAbI data, and test on bAbI+.


\subsubsection{Results: Prediction of \texttt{api-call}s}
\paragraph{BABBLE} Mere successful parsing of all the dialogues in the bAbI and bAbI+ datasets as shown above doesn't mean that the semantic representations compiled for the dialogues were in fact correct. To measure the semantic accuracy of the DS-TTR parser we programmatically checked that the correct slot values~-- those in the \texttt{api-call} annotations~-- were in fact present in the semantic representations produced by the parser for each dialogue (see Fig.~\ref{fig:subtype} for example semantic representations). We further checked that there is no other incorrect slot value present in these representations. The results showed that the parser has 100\% semantic accuracy on both bAbI and bAbI+\footnote{A helpful reviewer points out that the DS-TTR setup is a carefully tuned rule-based system, thus perhaps rendering these results trivial. But we note that the results here are not due to ad-hoc constructions of rules/lexicons, but due to the generality of the grammar model, and its attendant incremental, left-to-right properties; and that the same parser can be used in other domains. Furthermore, the ability to process self-corrections, restarts, etc. ``comes for free'', without the need to add or posit new machinery}. This result is not surprising, given that DS-TTR is a general model of incremental language processing, including phenomena such as self-corrections and restarts (see \newcite{Hough15} for details of the model). 



\paragraph{MEMN2N} Given just 1 to 5 training instances from bAbI as in the previous experiment, the mean \texttt{api-call} prediction accuracy of the \memnn\ model is \textbf{nearly 0 on both bAbI and bAbI+}. This is not at all unexpected, since prediction of the \texttt{api-call}s is a \emph{generative} process, unlike the prediction of system turns which can be done on a retrieval/look-up basis alone. For this, the model needs to observe the different word sequences that might determine each parameter (slot) value, and observe them with sufficient frequency and variation. This is unlike a semantic parser like DS-TTR, that produces semantic representations for the dialogues as a result of the structural, linguistic knowledge that it embodies.

Nevertheless, we were also interested in the general semantic robustness of the \memnn\ model, to the transformations in bAbI+, i.e.\ how well does the \memnn\ model interpret bAbI+ dialogues, when trained \emph{on the full bAbI dataset}? Does it then learn to generalise to (process) the bAbI+ dialogues with sufficient semantic accuracy?

Table~\ref{tab:sem-acc} shows that we can fully replicate the results reported in \newcite{babi}: the \memnn\ model can predict the \texttt{api-call}s with 100\% accuracy, when trained on the bAbI train-set and tested on the bAbI test-set. But when this same model is tested on bAbI+, the accuracy drops to a very poor 28\%, making any dialogue system built using this model unusable in the face of natural, spontaneous dialogue data. This is further discussed below.






\begin{table}
\begin{center}
\begin{tabular}{|c|c|}\hline
testing configuration&accuracy\\\hline \hline 
\textbf{\memnn\ on bAbI}&100\\\hline
\textbf{\memnn\ on bAbI+}&28\\\hline
\end{tabular}
\end{center} \caption{\texttt{api-call} prediction accuracies (\%) for the \memnn\ model trained on the bAbI trainset}\label{tab:sem-acc}
\end{table}

\section{Discussion}\label{discussion}
\vspace{-0.2cm}
\subsection{\babble}

The method described above has the following advantages over previous approaches to dialogue system development:

\textbf{--}  incremental (and thus more natural) language understanding, dialogue management,  and generation;


\textbf{--} ``end-to-end" method for task-based systems: no Dialogue Act annotations are required (i.e.\ reduced development time and effort); 

\textbf{--}  a complete dialogue system for a new   task can be automatically induced, using only `raw' data -- i.e.\ successful dialogue transcripts;

\textbf{--}  the MDP state and action spaces are automatically induced, rather than having to be designed by hand (as in prior work);

\textbf{--}  wide-coverage, task-based dialogue systems can be built from much smaller amounts of data as shown in section \ref{evaluation}
.


This final point bears further examination. As an empirically adequate model of incremental language processing in dialogue, the DS-TTR grammar is required to capture interactional variants such as question-answer pairs, over- and under-answering, self- and other-corrections, clarification, split-utterances, and ellipsis more generally.
As we showed in section \ref{evaluation}, even if  most of these structures are not  present in the training example(s), the resulting trained system is able to handle them, thus resulting in a very significant generalisation around the original data.

We also note that since we were in this instance interested in a direct comparison with {\memnns} over the bAbI \& bAbI+ datasets, we didn't exploit the power of Reinforcement Learning and exploration as we described above - as we have done before with other systems \cite{Kalatzis.etal16}. Therefore the generalisation results we report above for {\babble}\ follow entirely from the knowledge present within the grammar as a computational model of dialogue processing and contextual update, rather than this having been learned from data. Applying the full RL method described above would have meant that the system would actually discover many interactional and syntactic variations that are not present in bAbI, nor in bAbI+.

\vspace{-0.2cm}
\subsection{\memnn}
Even when trained on very few training instances, the \memnn\ model was able to predict system responses remarkably well. But results from Experiment 2 above showed that this was misleading: the \memnn s were making a drastic number of semantic mistakes when interpreting the dialogues, both in the bAbI and bAbI+ datasets. Even when trained on the full bAbI data-set, the model performed badly on bAbI+ in terms of semantic accuracy. We diagnose these results as follows:

\noindent\textbf{Problem complexity:} The first thing to notice is that in bAbI dialogue Task 1, the responses are highly predictable and stay constant regardless of the actual task details (slot values) up to the point of the final \texttt{api-call}s; and further, that the prediction of \texttt{api-call}s is a \emph{generative} process, unlike the prediction of the system turns, which is retrieval-based. This, in our view, explains the very large difference in \memnn\ performance across the two prediction tasks.\\

\noindent\textbf{Model robustness to the bAbI+ transformations:}.  The variations introduced in  bAbI+ are repetitions of both content and non-content words, as well additional incorrect slot values. The model was working in the same setup as \babble, therefore none of those variations could be treated as unknown tokens for either system. Although in the case of \memnn, some of the mixed-in words never appeared in the training data, and bAbI+ utterances were augmented significantly with those words -- so it was interesting to see how such untrained embeddings would affect the latent memory representations inside \memnn. The resulting performance suggests that there was no significant impact on {\memnn} from these variations as far as the predicting system responses was concerned. But the incorrect slot values introduced in self-corrections affect the system's task completion performance significantly, only appearing at the point of \texttt{api-call} predictions. 

We note also that none of our experiments in this paper involved training \memnn\ on bAbI+ data. There is a very interesting question here: is the \memnn\ model in principle able to learn to process the bAbI+ structures if it is in fact trained on it? And how much bAbI+ data would it require to do so? These issues are address in detail in \newcite{Shalyminov.etal17}.



\section{Conclusions}

Our main advances are in a) training end-to-end dialogue systems from small amounts of data, b)   incremental processing for   wider coverage of more natural everyday dialogues (e.g.\ containing self-repairs).

We compared our grammar-based approach to dialogue processing (DS-TTR) with a state-of-the-art, end-to-end response retrieval model (\memnns)  \cite{babi}, when training on small amounts of dialogue data.

Our experiments show that our model can process 74\% of the Facebook AI bAbI dataset even when trained on only 0.13\% of the data (5 dialogues). It can \textit{in addition} process 65\% of bAbI+, a corpus we created by systematically adding incremental dialogue phenomena such as restarts and self-corrections to bAbI. We find on the other hand that the \memnn\ model is not robust to the structures we introduced in bAbI+, even when trained on the full bAbI dataset.




\ignore{ 
We showed how end-to-end incremental dialogue systems can be automatically learned from a small number of example successful dialogues in a domain --  without the use of expensive and time-consuming  Dialogue Act annotations.
This method allows rapid domain transfer -- simply collect some example (successful) dialogues, and train. At present this is fully automated, and only requires checking that the DS lexicon covers the input data.
We evaluated the power of this method on a state-of-the-art dataset - the bAbI restuarant search task, and showed for example that  all the  bAbI test dialogues can be correctly processed by our system after training on only N dialogues.

Several questions remain -- for example regarding which classes of dialogues   this method can be used for. Here we have shown that standard task-oriented `slot-filling' dialogue systems can be automatically created in this way, but other dialogues involving deeper and more complex semantic relations ought to be amenable to the same treatment. 
For example, we are currently applying the same method to a visual language grounding and concept learning task. 
} 


\section*{Acknowledgements}
This research is  supported by the EPSRC, under grant number EP/M01553X/1 (BABBLE project\footnote{\url{https://sites.google.com/site/hwinteractionlab/babble}}).

\bibliography{babble,SG_B,all}
\bibliographystyle{emnlp_natbib}

 \end{document}